\documentclass{article}
\usepackage{spconf,amsmath,graphicx,hyperref}

\usepackage{booktabs}
\usepackage{cite}
\usepackage{amsmath,amssymb,amsfonts}
\usepackage{graphicx}
\usepackage{textcomp}
\usepackage{xcolor}
\usepackage{subfigure}
\usepackage{balance}

\usepackage{array}
\usepackage{booktabs}
\usepackage{url}
\usepackage{amsmath}
\usepackage{siunitx}
\usepackage{color}
\usepackage{multirow}
\usepackage{booktabs}
\usepackage{float}
\usepackage{bbding}
\usepackage{hyperref}

\usepackage{graphicx}
\usepackage{wrapfig}
\usepackage{amsmath}
\usepackage{amssymb}
\usepackage{algorithm}
\usepackage{algpseudocode}
\usepackage{booktabs}
\usepackage{multirow}
\usepackage{cite}
\usepackage{xcolor}
\usepackage{hyperref}
\hypersetup{
    colorlinks=true,
    linkcolor=blue,
    filecolor=magenta,      
    urlcolor=cyan,
    pdftitle={Overleaf Example},
    pdfpagemode=FullScreen,
}

\title{DPI: Exploiting Parameter Heterogeneity for Interference-Free Fine-Tuning}
\name{Xiaoyu Liu$^{1,*}$, Xiaoyu Guan$^{2,*}$, Di Liang, Xianjie Wu$^{3,\dagger}$\thanks{$^{\dagger}$Corresponding author,* Equal contribution}}
\address{$^{1}$	College of Science, Northeastern University, Boston, United States\\
$^{2}$ University of Florida, Gainesville, FL 32611, United States\\
$^{3}$Beijing Information Science and Technology University\\
liu.xiaoyu7@northeastern.edu,\quad\{guanxiaoyuyu, xianjie0822\}@gmail.com
}
\begin{document}
%
\maketitle
\begin{abstract}
Supervised fine-tuning (SFT) is a crucial step for adapting large language models (LLMs) to downstream tasks. However, conflicting objectives across heterogeneous SFT tasks often induce the “seesaw effect”: optimizing for one task may degrade performance on others, particularly when model parameters are updated indiscriminately. In this paper, we propose a principled approach to disentangle and isolate task-specific parameter regions, motivated by the hypothesis that parameter heterogeneity underlies cross-task interference. Specifically, we first independently fine-tune LLMs on diverse SFT tasks and identify each task’s core parameter region as the subset of parameters exhibiting the largest updates. Tasks with highly overlapping core parameter regions are merged for joint training, while disjoint tasks are organized into different stages. During multi-stage SFT, core parameters acquired in prior tasks are frozen, thereby preventing overwriting by subsequent tasks. 
To verify the effectiveness of our method, we conducted intensive experiments on multiple public datasets. The results showed that our dynamic parameter isolation strategy consistently reduced data conflicts and achieved consistent performance improvements compared to multi-stage and multi-task tuning baselines.
\end{abstract}

\section{Introduction}
\label{sec:introduction}

Large Language Models (LLMs) \cite{brown2020language,gao2025decorl} have shown remarkable capabilities across diverse natural language tasks. Supervised Fine-Tuning (SFT) \cite{wang2025not,wu2025progressive} is a key technique to adapt these models to specific applications and align them with human instructions. It involves training on curated input-output pairs to refine the model’s behavior.

\begin{figure}[ht]
\centering
\includegraphics[width=8.0cm]{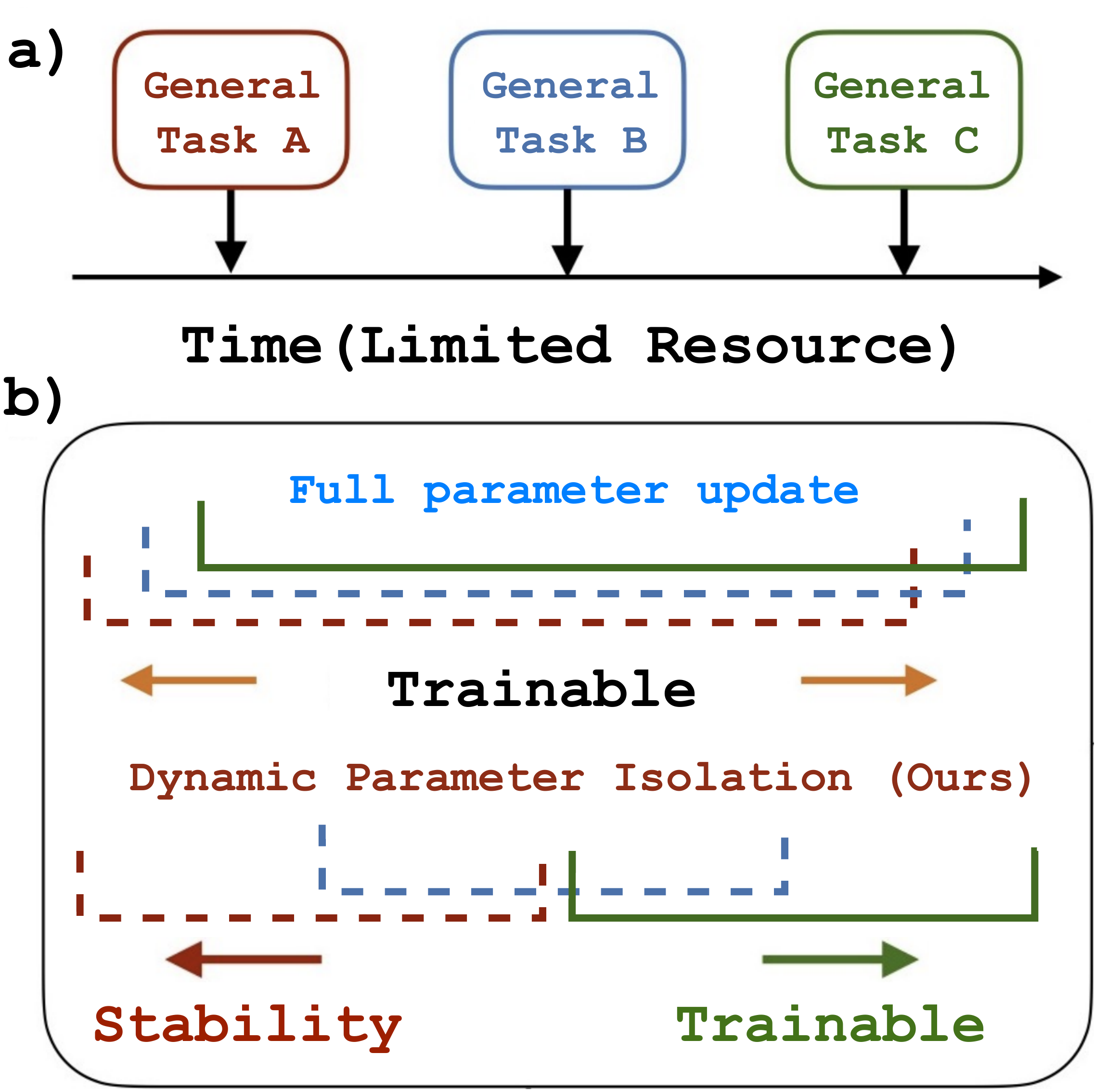}
\caption{Comparison of fine-tuning methods: (a) timeline of training data input; (b) difference between full-parameter SFT and dynamic parameter isolation fine-tuning.}
\label{fig:eurai}
\end{figure}

A major challenge arises when fine-tuning on a mixture of tasks, such as mathematical reasoning, creative writing, or coding, as these often involve conflicting objectives or reasoning mechanisms. Aggregating gradients across such tasks can lead to performance degradation, known as the "seesaw effect" \cite{yu2020gradient}, where gains in one task come at the expense of another.
Common solutions, like multi-stage training, group tasks heuristically and train them sequentially. While this reduces gradient conflict to some extent, it is a form of coarse-grained isolation and does not address parameter-level interference. It also increases the risk of catastrophic forgetting \cite{wu2025breaking}.
We hypothesize that the root cause of interference is \textit{parameter heterogeneity}: different tasks rely on distinct subsets of parameters within the LLM. Standard SFT, updating all parameters uniformly, fails to protect these specialized regions.

To address this, we propose \textbf{Dynamic Parameter Isolation (DPI)}, a novel SFT framework that identifies and isolates task-specific core parameters. DPI operates in two phases: first, it identifies core parameters for each task (or task group) by measuring update magnitudes during individual fine-tuning. Then, it designs a multi-stage training schedule: tasks with overlapping core parameters are trained together, while others are assigned to separate stages. Crucially, in each stage, core parameters from previous stages are frozen to avoid interference.
This approach minimizes destructive parameter updates and helps retain knowledge across tasks. We evaluate DPI on diverse public SFT datasets and show it consistently outperforms standard multi-task and multi-stage fine-tuning baselines.

In summary, our contributions are: (1) identifying parameter heterogeneity as a key factor in task interference; (2) proposing DPI, a framework for data-driven core parameter identification and dynamic isolation; (3) demonstrating through experiments that DPI reduces interference and forgetting, leading to stronger multi-task performance.

\section{Related Work}
\label{sec:related}

\noindent\textbf{Supervised Fine-Tuning and Instruction Tuning}  
SFT adapts pre-trained LLMs to downstream tasks via training on input-output pairs \cite{brown2020language,wu2025tablebench,liang2019adaptive,fei2022cqg}. Instruction tuning, a form of SFT, uses instruction-response data to improve generalization \cite{wei2021finetuned,liu2023time,wang2022dabert}. Standard SFT updates all parameters, which can cause task conflict and the "seesaw effect" when tasks have opposing objectives \cite{yu2020gradient,liu2023local,liang2019asynchronous,zheng2022robust}. Our method addresses this by selectively updating task-relevant parameters.

\noindent\textbf{Task Interference and Knowledge Retention}  
Task interference and catastrophic forgetting are key challenges in multi-task and sequential learning \cite{mccloskey1989catastrophic,ma2022searching,song2022improving}. Existing solutions include regularization, replay \cite{rolnick2019experience,xue2023dual}, gradient manipulation \cite{chen2018gradnorm,li2024local}, and modular approaches like adapters \cite{pfeiffer2020adapterfusion,xue2024question,liu2024resolving}. However, these methods often fail to address parameter-level contention in LLMs. DPI directly identifies and isolates task-specific core parameters based on update magnitudes, preserving knowledge by freezing these parameters.

\noindent\textbf{Multi-Stage Fine-Tuning and Dynamic Scheduling}  
Multi-stage fine-tuning heuristically groups tasks into stages to reduce gradient conflict \cite{wei2021finetuned}, but ignores parameter overlap, leading to forgetting. Multi-task fine-tuning trains tasks concurrently but struggles with gradient competition. Dynamic scheduling prioritizes tasks based on difficulty or similarity , yet operates at a data level. Our DPI performs parameter-level disentanglement, dynamically freezing task-specific regions during multi-stage training to minimize interference.

\noindent\textbf{Parameter Heterogeneity and Isolation}  
Parameter heterogeneity suggests that different parameters contribute unequally to tasks \cite{frankle2019winning,li2024comateformer}. Methods like adapters and LoRA exploit this by adding task-specific parameters \cite{houlsby2019bertadapter,dai2025hope,wu2025unleashing}, but they introduce overhead. Other work explores parameter reuse without explicit quantification \cite{liu2025stole,liu2025structural}. Our DPI method data-drivenly identifies task-specific core parameters without adding extra parameters, leveraging the natural heterogeneity in LLMs.

\section{Dynamic Parameter Isolation Tuning}
\label{sec:methodology}

This section presents the Dynamic Parameter Isolation (DPI) framework, a novel approach for supervised fine-tuning that mitigates task interference and catastrophic forgetting in multi-task learning scenarios. DPI operates on the principle of parameter heterogeneity within large language models, systematically identifying and protecting task-specific parameter regions through a three-stage process: core parameter identification, task grouping based on parameter similarity, and multi-stage training with dynamic freezing.

\subsection{Stage 1: Identifying Task-Specific Core Parameter Regions}

The foundation of DPI rests on the observation that different SFT tasks utilize distinct parameter subsets within LLMs. To quantitatively identify these specialized regions, we employ a probing procedure for each task $T_i \in \mathcal{T}$. Starting from the pre-trained checkpoint $\theta^{(0)}$, we perform a short fine-tuning session using only the task-specific dataset $\mathcal{D}i$:
\begin{equation}
\theta^{(i)} = \text{SFT}(\mathcal{M}(\theta^{(0)}), \mathcal{D}i, E{\text{probe}})
\end{equation}
where $E{\text{probe}}$ represents a limited number of training epochs sufficient to induce meaningful parameter shifts without full convergence.

We then compute the absolute parameter update magnitudes:
\begin{equation}
\Delta |\theta^{(i)}_j| = |\theta^{(i)}_j - \theta^{(0)}_j|, \quad \text{for } j = 1, ..., D
\end{equation}

The core parameter region $C_i$ for task $T_i$ is defined as the subset of parameters exhibiting the largest update magnitudes:
\begin{equation}
C_i = \text{arg topk}_{j \in {1..D}} (\Delta |\theta^{(i)}_j|, \lfloor p \cdot D / 100 \rfloor)
\end{equation}
where $p$ is a hyperparameter controlling the size of the core region, typically set to 1-5

\subsection{Stage 2: Task Grouping and Staging via Core Region Similarity}

To organize the training process, we group tasks based on the similarity of their identified core parameter regions. We quantify the overlap between core regions $C_i$ and $C_j$ using the Jaccard Index:
\begin{equation}
S(C_i, C_j) = \frac{|C_i \cap C_j|}{|C_i \cup C_j|} \in [0, 1]
\end{equation}

Tasks are grouped using a threshold-based clustering approach where $T_i$ and $T_j$ are considered related if:
\begin{equation}
T_i \sim T_j \iff S(C_i, C_j) \geq \tau
\end{equation}
with $\tau$ typically set between 0.2-0.4 based on validation performance. The final task groups $G_1, G_2, ..., G_K$ are formed by computing connected components of this similarity graph.

\subsection{Stage 3: Multi-Stage SFT with Dynamic Parameter Freezing}

The training proceeds sequentially through stages $k=1, ..., K$. At each stage, we protect parameters critical to previous tasks by constructing a frozen parameter set:
\begin{equation}
F_k = \bigcup_{l=1}^{k-1} \bigcup_{T_i \in G_l} C_i
\end{equation}

A binary mask $M_k$ is applied during optimization to prevent updates to frozen parameters while allowing updates to the remaining parameters:
\begin{equation}
\theta_{t+1} = \theta_t + \Delta \theta_t \odot M_k
\end{equation}
where $\odot$ denotes element-wise multiplication. This dynamic freezing mechanism ensures that knowledge acquired in earlier stages is preserved while allowing subsequent tasks to utilize the non-frozen parameter space for learning new capabilities.
The complete DPI algorithm provides an efficient implementation of this framework, enabling effective multi-task fine-tuning while minimizing destructive interference between competing objectives.

\begin{table*}[ht]
\centering
\label{tab:main_results_expanded}
\resizebox{\textwidth}{!}{
\renewcommand{\arraystretch}{0.9}
\begin{tabular}{llcccccc}
\toprule
Base Model & Method & GSM8K (Acc \%) & CodeAlpaca (CodeBLEU) & LogiQA (Acc \%) & Alpaca (GPT-4) & UltraChat (GPT-4) & Avg. Norm. Score \\
\midrule
\multirow{4}{*}{LLaMA-2-7B} & Full SFT (Multi-task) & 48.2 & 25.1 & 55.3 & 7.1 & 7.5 & 6.58 \\
& Multi-Stage (Random, K=3) & 49.5 & 24.8 & 56.0 & 7.3 & 7.6 & 6.70 \\
& Multi-Stage (Heuristic) & 50.1 & 25.5 & 56.8 & 7.0 & 7.4 & 6.75 \\
& \textbf{DPI (Ours, p=1\%, $\tau$=0.1)} & \textbf{53.4} & \textbf{27.1} & \textbf{59.2} & \textbf{7.6} & \textbf{7.9} & \textbf{7.18} \\
\midrule
\multirow{4}{*}{Mistral-8B} & Full SFT (Multi-task) & 46.5 & 24.0 & 53.8 & 6.9 & 7.3 & 6.37 \\
& Multi-Stage (Random, K=3) & 47.8 & 23.7 & 54.5 & 7.1 & 7.4 & 6.49 \\
& Multi-Stage (Heuristic) & 48.3 & 24.3 & 55.2 & 6.8 & 7.2 & 6.53 \\
& \textbf{DPI (Ours, p=1\%, $\tau$=0.1)} & \textbf{51.5} & \textbf{25.8} & \textbf{57.5} & \textbf{7.4} & \textbf{7.7} & \textbf{6.96} \\
\midrule
\multirow{4}{*}{Qwen1.5-7B} & Full SFT (Multi-task) & 49.8 & 26.0 & 56.5 & 7.3 & 7.7 & 6.79 \\
& Multi-Stage (Random, K=3) & 51.0 & 25.7 & 57.3 & 7.5 & 7.8 & 6.92 \\
& Multi-Stage (Heuristic) & 51.7 & 26.4 & 58.0 & 7.2 & 7.6 & 6.98 \\
& \textbf{DPI (Ours, p=1\%, $\tau$=0.1)} & \textbf{55.1} & \textbf{28.0} & \textbf{60.5} & \textbf{7.8} & \textbf{8.1} & \textbf{7.41} \\
\midrule
\multirow{4}{*}{Gemma-9B} & Full SFT (Multi-task) & 51.5 & 27.2 & 58.0 & 7.6 & 8.0 & 7.05 \\
& Multi-Stage (Random, K=3) & 52.8 & 26.9 & 58.9 & 7.8 & 8.1 & 7.19 \\
& Multi-Stage (Heuristic) & 53.5 & 27.6 & 59.7 & 7.5 & 7.9 & 7.26 \\
& \textbf{DPI (Ours, p=1\%, $\tau$=0.1)} & \textbf{57.0} & \textbf{29.3} & \textbf{62.4} & \textbf{8.1} & \textbf{8.4} & \textbf{7.70} \\
\bottomrule
\end{tabular}%
}
\caption{Main performance comparison of baselines on various SFT tasks. Higher scores indicate better performance. Best results for each task are shown in \textbf{bold}. The Avg. Norm. Score is computed by normalizing individual task scores to 10 scale.}
\end{table*}

\section{Experiments Setup}
\label{sec:experiments}


\noindent\textbf{Datasets and Baselines}
We evaluate on five diverse public datasets spanning mathematical reasoning (GSM8K, accuracy), code generation (CodeAlpaca, CodeBLEU), logical reasoning (LogiQA, accuracy) , and instruction following (Alpaca and UltraChat, GPT-4 score) . We report both task-specific metrics and a macro-average normalized score (0-10 scale) for unified comparison.
We compare against three standard SFT approaches: (1) \textit{Full Multi-task SFT}: uniform mixture of all datasets; (2) \textit{Multi-Stage SFT (Random)}: 3-stage sequential training with random task grouping; (3) \textit{Multi-Stage SFT (Heuristic)}: 2-stage sequential training with domain-based grouping.

\section{Main Results}
\label{sec:results}

The primary results comparing our proposed Dynamic Parameter Isolation with baseline approaches across multiple base models and SFT tasks are presented in Table \ref{tab:main_results_expanded}. We evaluate performance on five tasks (GSM8K, CodeAlpaca, LogiQA, Alpaca, UltraChat) and report the overall Average Normalized Score. Experiments were conducted using four foundation models: LLaMA-2-7B, Mistral-8B, Qwen1.5-7B, and Gemma-9B.

DPI consistently outperforms all baselines across tasks and models, achieving the highest scores on every individual task and the highest Average Normalized Score. Full SFT (Multi-task) performs worst due to task interference, while Multi-Stage SFT (Random or Heuristic) offers some improvement but is still outperformed by DPI. The heuristic variant provides only marginal gains over random staging, whereas DPI's data-driven core parameter identification and dynamic freezing deliver substantial benefits.

These results hold across different architectures and model sizes, indicating that DPI's parameter heterogeneity-aware mechanism generalizes beyond a specific model family. All DPI results were obtained with $p=1\%$ (core percentage) and $\tau=0.1$ (similarity threshold).

In summary, DPI effectively mitigates task conflict by protecting task-specific core parameters, leading to significant performance improvements across diverse tasks and base models compared to standard SFT.


\subsection{Ablation Studies}
\begin{figure}[ht]
\centering
\includegraphics[width=8.0cm]{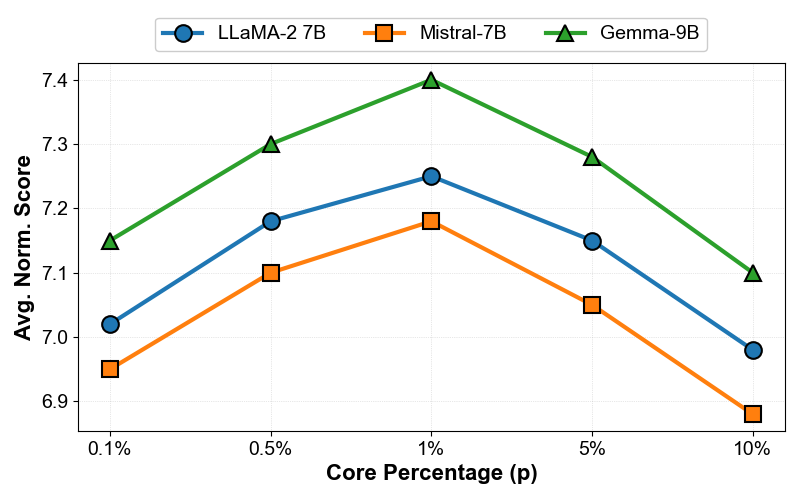}
\caption{Impact of core percentage ($p$).}
\label{fig:figure_p}
\end{figure}
We vary the percentage $p$ used to define core parameter regions. This hyperparameter determines the size of the parameter subset considered crucial for each task. We test values ranging from 0.1\% to 10\%. Results across different base models are shown in Figure \ref{fig:figure_p}. All runs use the similarity threshold $\tau=0.1$.
Performance generally peaks around $p=0.5\%$ to $p=1\%$ across the tested models . This suggests a relatively small fraction of parameters captures the most significant task-specific adaptations relevant for isolation. Using very small values (e.g., 0.1\%) might not identify a sufficiently comprehensive set of critical parameters, leading to suboptimal isolation. Conversely, using larger values (e.g., 5-10\%) might unnecessarily freeze parameters that could be beneficial for multiple task groups or could excessively hinder the model's plasticity, thereby degrading overall performance.



\section{Conclusion}
\label{sec:conclusion}
\label{sec:conclusion}

In this paper, we introduced Dynamic Parameter Isolation (DPI) Tuning, a principled framework for supervised fine-tuning (SFT) of large language models (LLMs) that mitigates task interference by identifying and isolating task-specific core parameter regions. By leveraging dynamic freezing during multi-stage fine-tuning, DPI preserves critical parameters for earlier tasks while enabling specialization for new ones. Extensive experiments on diverse datasets demonstrated DPI's effectiveness in addressing the "seesaw effect", reducing catastrophic forgetting, and consistently outperforming multi-task and multi-stage fine-tuning baselines. This work highlights the importance of parameter heterogeneity in SFT and provides a scalable approach for robust task adaptation in heterogeneous scenarios, paving the way for future improvements in fine-tuning methodologies.


\end{document}